# Automatic *Keyword* Extraction for Text Summarization: A Survey

Santosh Kumar Bharti[1], Korra Sathya Babu[2], and Sanjay Kumar Jena[3]

National Institute of Technology, Rourkela, Odisha 769008 India e-mail: {[1]sbharti1984, [2]prof.ksb}@gmail.com, [3]skjena@nitrkl.ac.in

## ABSTRACT

In recent times, data is growing rapidly in every domain such as news, social media, banking, education, etc. Due to the excessiveness of data, there is a need of automatic summarizer which will be capable to summarize the data especially textual data in original document without losing any critical purposes. Text summarization is emerged as an important research area in recent past. In this regard, review of existing work on text summarization process is useful for carrying out further research. In this paper, recent literature on automatic keyword extraction and text summarization are presented since text summarization process is highly depend on keyword extraction. This literature includes the discussion about different methodology used for keyword extraction and text summarization. It also discusses about different databases used for text summarization in several domains along with evaluation matrices. Finally, it discusses briefly about issues and research challenges faced by researchers along with future direction.

*Keywords:*
*Abstractive summary, extractive summary, Keyword Extraction, Natural language processing, Text Summarization.*

## 1. INTRODUCTION

In the era of internet, plethora of online information are freely available for readers in the form of e-Newspapers, journal articles, technical reports, transcription dialogues etc. There are huge number of documents available in above digital media and extracting only relevant information from all these media is a tedious job for the individuals in stipulated time. There is a need for an automated system that can extract only relevant information from these data sources. To achieve this, one need to mine the text from the documents. Text mining is the process of extracting large quantities of text to derive high-quality information. Text mining deploys some of the techniques of natural language processing (NLP) such as parts-of-speech (POS) tagging, parsing, N-grams, tokenization, etc., to perform the text analysis. It includes tasks like automatic keyword extraction and text summarization.

Automatic keyword extraction is the process of selecting words and phrases from the text document that can at best project the core sentiment of the document without any human intervention depending on the model [1]. The target of automatic keyword extraction is the application of the power and speed of current computation abilities to the problem of access and recovery, stressing upon information organization without the added costs of human annotators.

Summarization is a process where the most salient features of a text are extracted and compiled into a short abstract of the original document [2]. According to Mani and Maybury [3], text summarization is the process of distilling the most important information from a text to produce an abridged version for a particular task and user. Summaries are usually around 17\% of the original text and yet contain everything that could have been learned from reading the original article [4]. In the wake of big data analysis, summarization is an efficient and powerful technique to give a glimpse of the whole data. The text summarization can be achieved in two ways namely, abstractive summary and extractive summary. The abstractive summary is a topic under tremendous research; however, no standard algorithm has been achieved yet. These summaries are derived from learning what was expressed in the article and then converting it into a form expressed by the computer. It resembles how a human would summarize an article after reading it. Whereas, extractive summary extract details from the original article itself and present it to the reader.

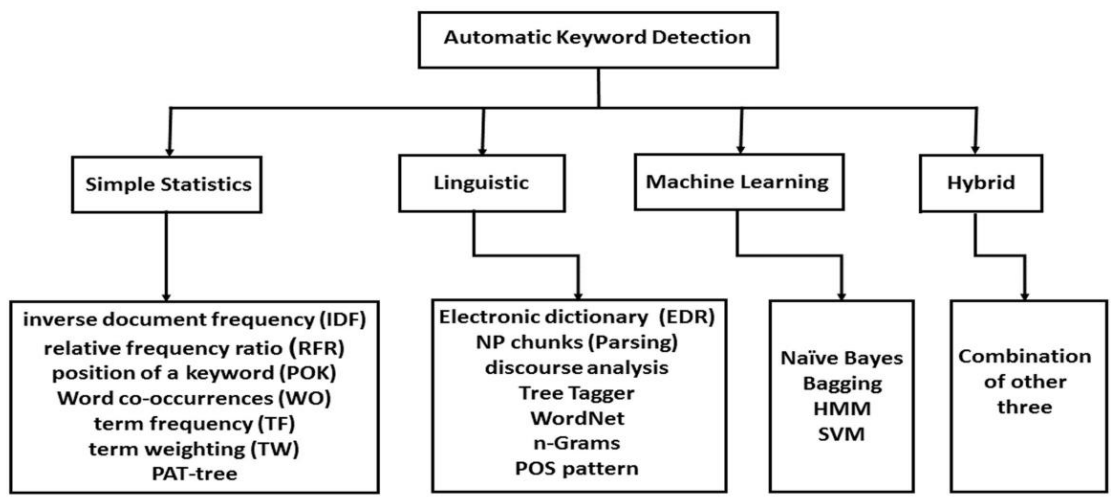

**Figure 1: Classification of automatic keyword extraction on the basis of approaches used in existing literature**

In this paper, reviewed the recent literature on automatic keyword extraction and text summarization. The valuable keywords extraction is the primary phase of text summarization. Therefore, in this literature, we focused on both the techniques. In keyword extraction, the literature discussed about different methodologies used for keyword extraction process and what algorithms used under each methodology as shown in Figure 1. It also discussed about different domains in which keyword extraction algorithms applied. Similarly, in the process of text summarization, literature covers all the possible process of text summarization such as document types (single or multi), summary types (generic or query based), techniques (supervised or unsupervised), characteristics of summary (abstractive or extractive), etc. Further, it includes all the possible methodologies for text summarization as shown in Figure 3.

This literature also discusses about different databases used for text summarization such as DUC, TAC, MEDLINE, etc. along with differnt evaluation matrices such as precision, recall, and ROUGE series. Finally, it discusses briefly about issues and research challenges faced by researchers followed by future direction of text summarization.

The rest of this paper is organized as follows. Section 2 presents a review on automatic keyword extraction Section 3 presents a review on text summarization process. A review on automatic text summarization methodologies are given in Section 4. The details about different databases are described in Section 5. Section 6 explains about evaluation matrices for text summarization. Finally, the conclusion with future direction is drawn in Section 7.

## 2   AUTOMATIC   KEYWORD   EXTRACTION:   A REVIEW

On the premise of past work done towards automatic keyword extraction from the text for its summarization, extraction systems can be classified into four classes, namely, simple statistical approach, linguistics approach, machine learning approach, and hybrid approaches [1] as soon in Figure 1.

### 2.1   Simple Statistical Approach

These strategies are rough, simplistic and have a tendency to have no training sets. They concentrate on statistics got from non-linguistic features of the document, for example, the position of a word inside the document, the term frequency, and inverse document frequency. These insights are later used to build up a list of keywords. Cohen [15], utilized n-gram statistical data to discover the keyword inside the document automatically. Other techniques in- side this class incorporate word frequency, term frequency (TF) [16] or term frequency-inverse document frequency (TF-IDF) [17], word co-occurrences [18], and PAT-tree [19]. The most essential of them is term frequency. In these strategies, the frequency of occurrence is the main criteria that choose whether a word is a keyword or not. It is extremely unrefined and tends to give very unseemly results. An improvement of this strategy is the TF-IDF, which also takes the frequency of occurrence of a word as the model to choose a keyword or not. Similarly, word co-occurrence methods manage statistical information about the number of times a word has happened and the number of times it has happened with another word. This statistical information is then used to compute support and confidence of the words. Apriori technique is then used to infer the keywords.

### 2.2   Linguistics Approach

This approach utilizes the linguistic features of the words for keyword detection and extraction in text documents. It incorporates the lexical analysis [20], syntactic analysis [21], discourse analysis [22], etc. The resources used for lexical analysis are an electronic dictionary, tree tagger, WordNet, n-grams, POS pattern, etc. Similarly, noun phrase (NP), chunks (Parsing) are used as resources for syntactic analysis.

### 2.3   Machine Learning Approach

Keyword extraction can also be seen as a learning problem. This approach requires manually annotated training data and

**Table 1: Previous studies on automatic keyword extraction**

| Study | Types of Approach | | | | Domain Types | | | | | | |
|---|---|---|---|---|---|---|---|---|---|---|---|
| | T1 | T2 | T3 | T4 | D1 | D2 | D3 | D4 | D5 | D6 | D7 |
| Dennis *et al.*[29], 1967 | | √ | | | | | √ | | | | |
| Salton *et al.*[30], 1991 | | √ | | | | | | | | √ | |
| Cohen *et al.*[15], 1995 | √ | | | | | √ | | | | | |
| Chien *et al.*[19], 1997 | √ | | | | | √ | | | | | |
| Salton *et al.*[22], 1997 | | √ | | | | | | | | √ | |
| Ohsawa *et al.*[31], 1998 | √ | | | | | √ | | | | | |
| Hovy *et al.*[2], 1998 | | √ | | | | | √ | | | | |
| Fukumoto *et al.*[32], 1998 | √ | | | | √ | | √ | | | √ | |
| Mani *et al.*[3], 1999 | | √ | | | | | | | | | |
| Witten *et al.*[26], 1999 | | | √ | | | | | √ | | | |
| Frank *et al.*[25], 1999 | | | √ | | | √ | | √ | | | |
| Barzilay *et al.*[20], 1999 | | √ | | | | | | | | | |
| Turney *et al.*[27], 1999 | | | √ | | | √ | | | | | |
| Conroy *et al.*[23], 2001 | | | √ | | | | | √ | | | |
| Humphreys *et al.*[28], 2002 | | √ | | √ | | | | | | | √ |
| Hulth et al.[21], 2003 | | √ | √ | | | √ | | √ | | | |
| Ramos *et al.*[17], 2003 | √ | | | | | | | | | | |
| Matsuo *et al.*[18], 2004 | √ | | | | | | | | | | |
| Erkan *et al.*[4], 2004 | | | √ | | | | | | | | |
| Van *et al.*[6], 2004 | | | √ | | | | | | √ | | |
| Mihalcea *et al.*[33], 2004 | | | √ | | | | √ | | | | |
| Zhang *et al.*[24], 2006 | | | √ | | | √ | | | | | |
| Ercan *et al.*[8], 2007 | | | √ | | | √ | | | | | |
| Litvak *et al.*[9], 2008 | | | √ | | | | | | | | √ |
| Zhang *et al.*[1], 2008 | | | √ | | | √ | | | | | |
| Thomas *et al.*[5], 2016 | √ | √ | | √ | | | √ | | | | |

**Table 2: Types of approach and domains used in keyword extraction**

| | Types of Approach |
|---|---|
| T1 | Simple Statistics (SS) |
| T2 | Linguistics (L) |
| T3 | Machine Learning (ML) |
| T4 | Hybrid (H) |
| | **Types of Domain** |
| D1 | Radio News (RN) |
| D2 | Journal Articles (JA) |
| D3 | Newspaper Articles (NA) |
| D4 | Technical Reports (TR) |
| D5 | Transcription Dialogues (TD) |
| D6 | Encyclopedia Article (EA) |
| D7 | Web Pages (WP) |

training models. Hidden Markov model [23], support vector machine (SVM) [24], naive Bayes (NB) [25], bagging [21], etc. are commonly used training models in these approaches. In the second phase, the document whose keywords are to be extracted is given as inputs to the model, which then extracts the keywords that best fit the model's training. One of the most famous algorithms in this approach is the keyword extraction algorithm (KEA) [26]. In this approach, the article is first converted into a graph where each word is treated as a node, and whenever two words appear in the same sentence, the nodes are connected with an edge for each time they appear together. Then the number of edges connecting the vertices are converted into scores and are clustered accordingly. The cluster heads are treated as keywords. Bayesian algorithms use the Bayes classifier to classify the word into two categories: keyword or not a keyword depending on how it is trained. GenEx [27] is another tool in this approach.

### 2.4 Hybrid Approach

These approaches combine the above two methods or use heuristics, such as position, length, layout feature of the words, HTML tags around the words, etc. [28]. These algorithms are designed to take the best features from above mentioned approaches.

Based on the classification shown in Figure 1, we bring a consolidated summary of previous studies on automatic keyword extraction and is shown in Table 1. It discusses the approaches that are used for keyword extraction and various domains of dataset in which experiments are performed as shown in Table 2.

## 3. TEXT SUMMARIZATION PROCESS: A REVIEW

Based on the literature, text summarization process can be characterized into five types, namely, based on the number of the document, based on summary usage, based on techniques, based on characteristics of summary as text and based on levels of linguistics process [1] as shown in Figure 2.

### 3.1 Single Document Text Summarization

In single document text summarization, it takes a single document as an input to perform summarization and produce a single output document [5][34][35][36][37]. Thomas *et al.* [5] designed a system for automatic keyword extraction for text summarization in single document e-Newspaper article. Marcu *et al.* [35] developed a discourse-based summarizer that determines adequacy for summarizing texts for discourse-based methods in the domain of single news articles.

### 3.2 Multiple Document Text Summarization

In multiple documents text summarization, it takes numerous documents as an input to perform summarization and deliver a single output document [14][38][39][40][41][42][43][44]. Mirroshandel *et al.* [44] presents two different algorithms towards temporal relation based keyword extraction and text summarization in multi-document. The first algorithm was a weakly supervised machine learning approach for classification of temporal relations between events and the second algorithm was expectation maximization (EM) based unsupervised learning approach for temporal relation extraction. Min *et al.* [40] used the information which is common to document sets belonging to a common category to improve the quality of automatically extracted content in multi-document summaries.

### 3.3 Query-based Text Summarization

In this summarization technique, a particular portion is utilized to extract the essential keyword from input document to make the summary of corresponding document [11][37][48][49][50][51]. Fisher *et al.* [50] developed a query-based summarization system that uses a log-linear model to classify each word in a sentence. It exploits the property of sentence ranking methods in which they consider neural query ranking and query-focused ranking. Dong *et al.* [51] developed a query-based summarization that uses document ranking, time-sensitive queries and ranks recency sensitive queries as the features for text summarization.

### 3.4 Extractive Text Summarization

In this procedure, summarizer discovers more critical information (either words or sentences) from input document to make the summary of the corresponding document [2][5][52][35][53][54][55][65][75][39][40]. In this process, it uses statistical and linguistic features of the sentences to decide the most relevant sentences in the given input document. Thomas *et al.* [5] designed a hybrid model based extractive summarizer using machine learning and simple statistical method for keyword extraction from e-Newspaper article. Min *et al.* [40] used freely available, open-source extractive summarization system, called SWING to summarize the text in multi-document. They used information which is common to document sets belonging to a common category as a feature and encapsulated the concept of category-specific importance (CSI). They showed that CSI is a valuable metric to aid sentence selection in extractive summarization tasks. Marcu *et al.* [35] developed a discourse-based extractive summarizer that uses the rhetorical parsing algorithm to determine discourse structure of the text of given input, determine partial ordering on the elementary and parenthetical units of the text. Erkan *et al.* [65] developed an extractive summarization environment. It consists of three steps: feature extractor, the feature vector, and reranker.

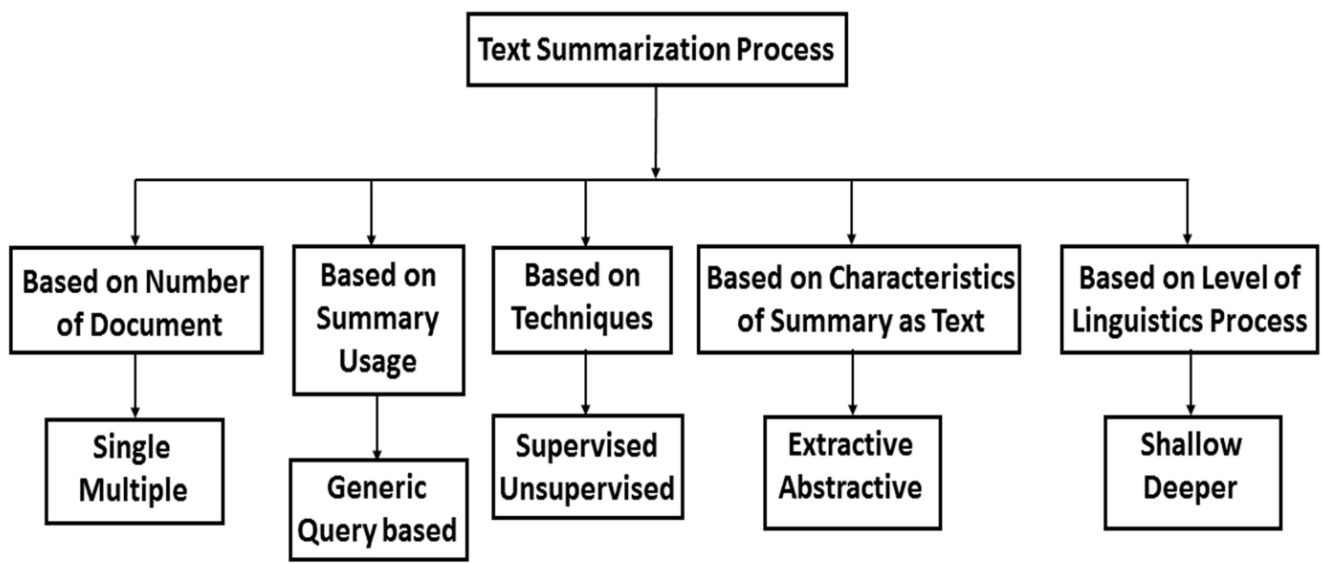

**Figure 2: Characterization of the text summarization process**

Features are Centroid, Position, Length Cutoff, SimWithFirst, LexPageRank, and QueryPhraseMatch. Alguliev *et al.* [39] developed an unsupervised learning based extractive summarizer that optimizes three properties: relevance, redundancy, and length. It split documents into sentences and select salient sentences from the document. Aramaki *et al.* [75] destined a supervised learning based extractive text summarizer that identifies the negative event and it also investigates what kind of information is helpful for negative event identification. An SVM classifier is used to distinguish negative events from other events.

### 3.5 Abstractive Text Summarization

In this procedure, a machine needs to comprehend the idea of all the input documents and then deliver summary with its particular sentences [34][37][52][56][57][58]. It uses linguistic methods to examine and interpret the text and then to find the new concepts and expressions to best describe it by generating a new shorter text that conveys the most important information from the original text document. Brandow *et al.* [56] developed an abstractive summarization system that analyses the statistical corpus and extracts the signature words from the corpus. Then it assigns the weight for all the signature words. Based on the extracted signature words, they assign the weight to the sentences and select few top weighted sentences as the summary. Daume *et al.* [37] developed an abstractive summarization system that maps all the documents into database-like representation. Further, it classifies into four categories: a single person, single event, multiple event, and natural disaster. It generates a short headline using a set of predefined templates. It generates summaries by extracting sentences from the database.

### 3.6 Supervised Learning Based Text Summarization

This type of learning techniques used labeled dataset for training [5][12][38][40][44][50][73][75]. Thomas et al. [5] designed a system for automatic keyword extraction for text summarization using hidden Markov model. The learning process was supervised, it used human annotated keyword set to train the model. Mirroshandel *et al.* [44] used a set of labeled dataset to train the system for the classification of temporal relations between events. Aramaki et al. [75]

destined a supervised learning based extractive text summarizer that identifies the negative event and also investigates what kind of information is helpful for negative event identification. An SVM classifier is used to distinguish negative events from other events.

### 3.7 Unsupervised Learning Based Text Summarization

In this technique, there are no predefined guidelines available at the time of training [13][38][39][44][65]. Mirroshandel *et al.* [44] proposed a method for temporal relation extraction based on the Expectation-Maximization (EM) algorithm. Within EM, they used different techniques such as a greedy best-first search and integer linear programming for temporal inconsistency removal. The EM-based approach was a fully unsupervised temporal relation based extraction for text summarization. Alguliev *et al.* [39] developed an unsupervised learning based extractive summarizer that optimizes three properties: relevance, redundancy, and length. It split documents into sentences and select salient sentences from the document.

### 2. TEXT SUMMARIZATION APPROACH: A REVIEW

Based on the literature, text summarization approaches can be classified into five types, namely, statistical based, machine learning based, coherent based, graph based, algebraic based as shown in Figure 3.

### 4.1 Statistical Based Approach

This approach is very simple and crude often used for keyword extraction from the documents. There is no predefined dataset required for this approach. To extract the keywords from documents it uses several statistical features of the document such as, term or word frequency (TF), Term Frequency-inverse document frequency (TF-IDF), position of keyword (POK), etc. as mentioned in Figures 1 and 3. These statistical features are used for text summarization [5][90][91][92].

### 4.2 Machine learning Based Approach

Machine learning is a feature dependent approach we one need annotated dataset to trained the models. There are several popular machine learning approaches namely, Nave Bayes (NB) [93][94][95], decision trees (DTs) [96][97], Hidden Markov Model (HMM) [98][99][100], Maximum Entropy (ME) [101][102][103][104][105], Neural Network (NN) [106][107][108], Support Vector Machine (SVM) [109][110][111] etc. used for text summarization.

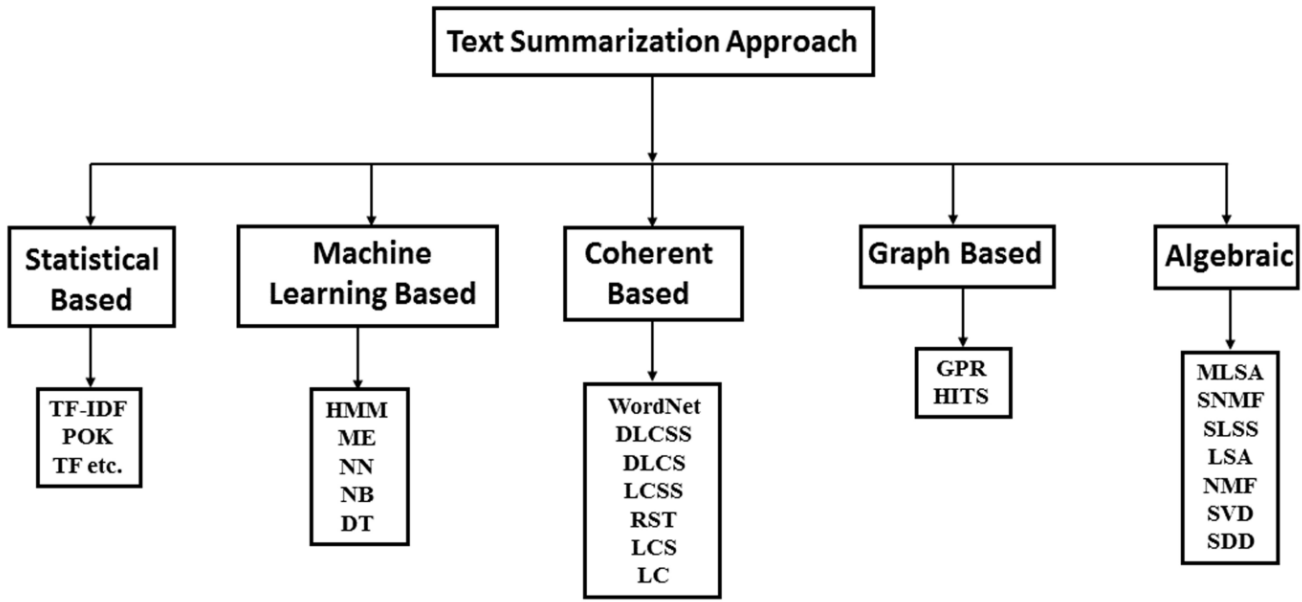

**Figure 3:  Classification of Text Summarization Approaches**

**Table 3: List of Abbreviations used in Classification of Text Summarization Approaches**

|         | List of Abbreviations |
|---------|------------------------|
| TF | Term Frequency |
| IF-IDF | Term Frequency-inverse document frequency |
| POK | Position of a Keyword |
| HMM | Hidden Markov Model |
| DT | Decision Trees |
| ME | Maximum Entropy |
| NN | Neural Networks |
| NB | Naïve Bayes |
| DLCSS | Direct lexical chain span score |
| DLCS | Direct lexical chain score |
| LCSS | Lexical chain span score |
| LCS | Lexical chain score |
| RST | Rhetorical Structure Theory |
| LC | Lexical chain |
| GPR | Google's Pagerank |
| HITS | Hyperlinked Induced Topic Search |
| MLSA | Meta Latent Semantic Analysis |
| SNMF | Symmetric nonnegative matrix |
| SLSS | Sentence level semantic analysis |
| LSA | Latent Semantic Analysis |
| NMF | Non-Negative Matrix factorization |
| SVD | Singular Value Decomposition |
| SDD | Semi-Discrete Decomposition |

### 4.3  Coherent Based Approach

A coherent based approach basically deals with the cohesion relations among the words. Cohesion relations among elements in a text: reference, ellipsis, substitution, conjunction, and lexical cohesion [112]. Lexical chain (LC) [113], WordNet (WN) [8][114], lexical chain score of a word (LCS) [113], direct lexical chain score of a word (DLCS) [113], lexical chain span score of a word (LCSS) [113], direct lexical chain span score of a word (DLCSS) [113], Rhetorical Structure Theory (RST) [115][116][9].

### 4.4  Graph Based Approach

There are two popular graph-based approaches used for text summarization namely, Hyperlinked Induced Topic Search (HITS) [117][118] and Google's PageRank (GPR) [117][33][119][120].

### 4.5  Algebraic Approach

In this approach, one use algebraic theories namely, matrix, transpose of matrix, Eigen vectors, etc. There are many algorithms used for text summarization using algebraic approach such as Latent Semantic Analysis (LSA) [121][122][123], Meta Latent Semantic Analysis (MLSA) [124][125], Symmetric nonnegative matrix factorization (SNMF) [126], Sentence level semantic analysis (SLSS) [126], Non-Negative Matrix factorization (NMF) [127], Singular Value Decomposition (SVD) [128], Semi-Discrete Decomposition (SDD) [129].

### 3. DATABASES: A REVIEW

In the literature, we observed that, there are seven types of databases used for text summarization process namely, document understanding workshop (DUC), MEDLINE, Text Analysis Conference (TAC), Computational Linguistics Scientific Document Summarization Shared Task Corpus (CL-SciSumm), TIPSTER Text Summarization Evaluation Conference (SUMMAC), Topic Detection and Tracking (TDT). The DUC is the international conference for performance evaluation in the area of text summarization. This dataset is composed of 50 topics and 25 documents relevant to each topic from the AQUAINT corpus for query-relevant multi-document summarization [21]. MEDLINE/PubMed dataset is a baseline repository that links 19 million articles to with http://dx.doi.org/ article identifiers and http://crossref.org/ with journal identifiers [130] and it contains abstracts from more than 3500 journals [131]. MEDLINE also provides keyword searches and returns abstracts that contain the keywords. The TAC is the international conference for performance evaluation in the area of subparts of Natural Language Processing such as text summarization. Usually, TAC- 2009, 2010 and 2011 conference dataset used for text summarization in past. The CL-SciSumm Shared Task is run off the CL-SciSumm corpus, and comprises three sub-tasks in automatic research paper summarization on a new corpus of research papers. A training corpus of twenty topics and a test corpus of ten topics were released. The topics comprised of ACL Computational Linguistics research papers, and their citing papers and three output summaries each. The three output summaries comprise: the traditional self-summary of the paper (the abstract), the community summary (the collection of citation sentences 'citances') and a human summary written by a trained annotator [132]. TIPSTER is a corpus of 183 documents from the Computation and Language (cmp-lg) collection has been marked up in xml and made available as a general resource to the information retrieval, extraction, and summarization communities.

## 4. 6. PERFORMANCE EVALUATION MEASURE: A REVIEW

In the field of text summarization, performance evaluation measure can be classified into two categories, namely, intrinsic and extrinsic. The intrinsic evaluation judges the quality of the summary directly based on analysis in terms of some set of norms whereas, extrinsic evaluation judges the quality of the summary based on the how it affects the completion of some other task. The complete structure of classification about evaluation measure for text summarization is given in Figure 4. To compare the performances, one used the ROUGE evaluation software package, which compares various summary results from several summarization methods with summaries generated by humans. ROUGE has been applied by the Document Understanding Conference (DUC) for performance evaluation. ROUGE includes five automatic evaluation methods, ROUGEN, ROUGE-L, ROUGE-W, ROUGE-S, and ROUGE-SU [24]. Each method estimates recall, precision, and f-measure between experts' reference summaries and candidate summaries of the proposed system. ROUGE-N uses the n-gram recall between a candidate summary and a set of reference summaries.

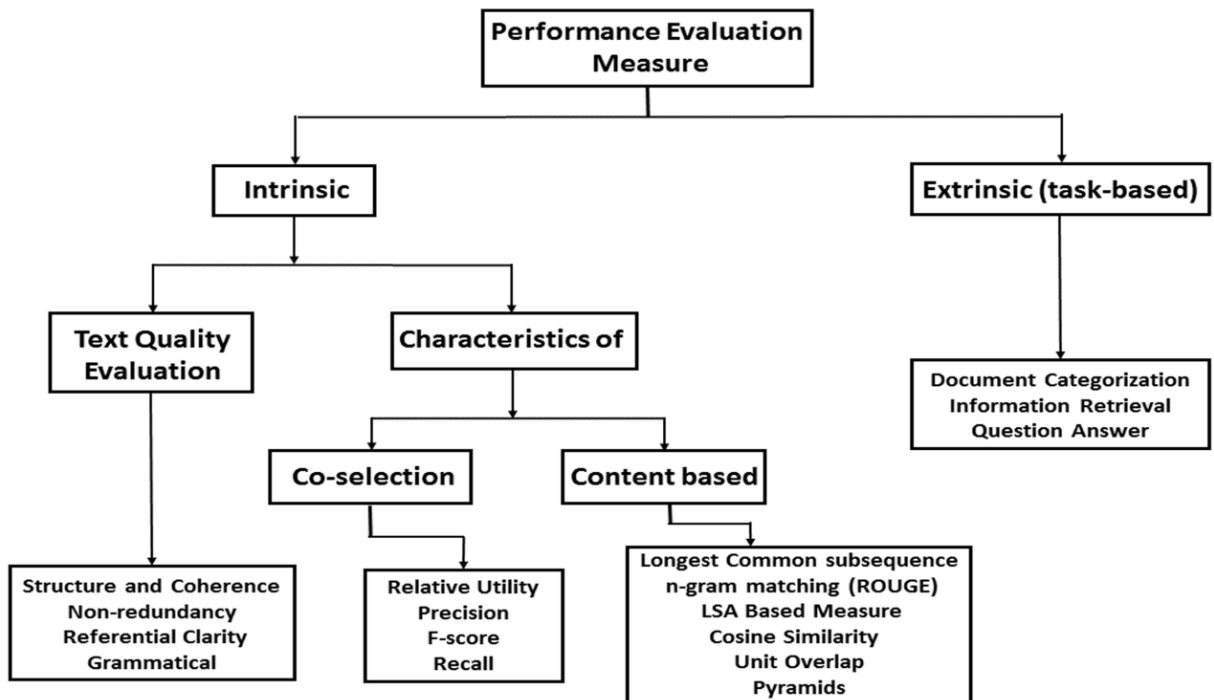

**Figure 4: Classification of Performance Evaluation Measure**

**Table 4: Previous studies on automatic text summarization**

| Study | TOA | | | | | DT | | TOSU | | COS | | DBU | | | | MAT | |
|---|---|---|---|---|---|---|---|---|---|---|---|---|---|---|---|---|---|
| | A1 | A2 | A3 | A4 | A5 | D1 | D2 | S1 | S2 | C1 | C2 | X1 | X2 | X3 | X4 | M1 | M2 |
| Pollock *et al.*[34], 1975 | | | | | √ | √ | | | | | √ | | | | | | |
| Brandow *et al.*[56], 1995 | | | | | √ | | √ | | | | √ | | | | | | √ |
| Hovy *et al.*[2], 1998 | | | | | √ | | √ | | | √ | | | | | | | √ |
| Aone *et al.*[45], 1998 | | | | | √ | √ | | √ | | √ | | | | | | | √ |
| Radev *et al.*[52], 1998 | | | | | √ | | √ | | | √ | √ | | | | | | |
| Marcu *et al.*[35], 1999 | | | | | √ | √ | | | | √ | | | | | | | √ |
| Barzilay *et al.*[57], 1999 | | | | | √ | | √ | | | | √ | | | | | | √ |
| Chen *et al.*[53], 2000 | | | | √ | | | √ | | | √ | | | | | | | √ |
| Radev *et al.*[54], 2001a | | | | | | √ | √ | | | √ | | | | | | | |
| Radev *et al.*[55], 2001b | | | | √ | | | √ | | | √ | | | | | | | |
| Radev *et al.*[46], 2001c | √ | | | | | | √ | √ | | √ | | | | | | | |
| Lin *et al.*[59], 2002 | | | | √ | | | √ | | | √ | | | | | | | √ |
| McKeown *et al.*[60], 2002 | | | | √ | | | √ | | | √ | | | | | | | |
| Daumé *et al.*[58], 2002 | | | | | √ | | √ | | | √ | √ | | | | | | |
| Harabagiu *et al.*[36], 2002 | | | | | √ | √ | √ | | | √ | √ | | | | | | √ |
| Saggion *et al.*[61], 2002 | | | | | √ | | √ | | | | √ | | | | | | √ |
| Saggion *et al.*[37], 2003 | | | | | √ | √ | | √ | √ | √ | | | | | | | √ |
| Chali *et al.*[62], 2003 | | | | √ | | √ | √ | | | √ | | | | | | | |
| Copeck *et al.*[63], 2003 | | | | | √ | √ | | | | √ | | | | | | | |
| Alfonseca *et al.*[64], 2003 | | | | | √ | √ | | | | √ | | | | | | √ | |
| Erkan *et al.*[65], 2004 | | | | √ | | | √ | | | √ | | | | | | √ | |
| Filatova *et al.*[10], 2004 | | | | | √ | | √ | | | √ | | | | | | √ | |
| Nobata *et al.*[66], 2004 | | | | | √ | | √ | | | √ | | | | | | | |
| Conroy *et al.*[11], 2005 | √ | | | | | | √ | | √ | √ | | | | | | √ | |
| Farzindar *et al.*[48], 2005 | √ | | | | | | √ | | √ | √ | | | | | | √ | |
| Witte *et al.*[67], 2005 | | | | | √ | | √ | | | √ | | | | | | √ | |
| Witte *et al.*[68], 2006 | | | | √ | | | √ | | | √ | | | | | | √ | |
| He *et al.*[69], 2006 | | | | | √ | | √ | | | √ | | | | | | √ | |
| Witte *et al.*[13], 2007 | | | | √ | | | √ | | | √ | | | | | | √ | |
| Fuentes *et al.*[49], 2007 | √ | | | | | √ | | | √ | √ | | | | | | | |
| Dunlavy *et al.*[70], 2007 | | | | | √ | √ | | | | √ | | | | | | √ | |
| Gotti *et al.*[71], 2007 | | | | | √ | | √ | | | √ | | | | | | √ | |
| Svore *et al.*[72], 2007 | √ | | | | | | √ | | | √ | | | | | | √ | |
| Schilder *et al.*[12], 2008 | √ | | | | | | √ | | | √ | | | | | | √ | |
| Liu et al.[73], 2008 | √ | | | | | | √ | | | √ | | | | | | √ | |
| Zhang *et al.*[74], 2008 | | | | | √ | | √ | | | √ | | | | | | √ | |
| Aramaki *et al.*[75], 2009 | √ | | | | | | √ | | | √ | | | | | | | √ |
| Fisher *et al.*[50], 2009 | √ | | | | | | √ | | √ | √ | | | | | | √ | |
| Hachey *et al.*[47], 2009 | | | | | √ | | √ | √ | | √ | | | | | | √ | |
| Wei *et al.*[76], 2010 | | | | | √ | | √ | | | √ | | | | | | | √ |
| Dong *et al.*[51], 2010 | | | | | √ | | √ | | √ | | | | | | | | |
| Shi *et al.*[38], 2010 | | | | | √ | | √ | | | √ | | | | | | | |
| Park *et al.*[87], 2010 | | | | √ | | | √ | | √ | √ | | | | | | √ | |
| Archambault *et al.*[14], 2011 | | | | | √ | | √ | | | | | | | | | | |
| Genest *et al.*[41], 2011 | | | | | | | √ | | | √ | | | | | | | |
| Tsarev *et al.*[42], 2011 | √ | | | √ | | | √ | | | √ | | | | | | √ | |
| Alguliev *et al.*[39], 2011 | | | | √ | | √ | √ | | | √ | | | | | | | |
| Chandra *et al.*[90], 2011 | √ | | | | | | | | | | | | | | | | |
| Mirroshandel *et al.*[44], 2012 | √ | | | √ | | | √ | | | √ | | | | | | | |
| Min *et al.*[40], 2012 | | | | | | | √ | √ | | √ | | | | | | √ | |
| De Melo *et al.*[43], 2012 | | | | | √ | | √ | | | √ | | | | | | | √ |
| Thomas *et al.*[5], 2012 | √ | | | | | √ | | | | √ | | | | | | | √ |

**Table 5: Approaches, documents, summary usage, characteristics of summary and metrics used in text summarization**

| | Types of Approach (TOA) |
|---|---|
| A1 | Statistical Approach |
| A2 | Machine Learning Approach |
| A3 | Coherent Based Approach |
| A4 | Graph Based Approach |
| A5 | Algebraic Approach |
| | Document Type (DT) |
| D1 | Single document |
| D2 | Multiple document |
| | Types of Summary Usage (TOSU) |
| S1 | Generic |
| S2 | Query based |
| | Characteristics of Summary (COS) |
| C1 | Extractive |
| C2 | Abstractive |
| | Databases Used (DBU) |
| X1 | DUC – 2001, 2002, 2003, 2004, 2005, 2006, 2007 |
| X2 | MEDLINE |
| X3 | TAC – 2009, 2010, 2011 |
| X4 | Others |
| | Performance Evaluation Metrics |
| M1 | ROUGE 1, ROUGE 2, ROUGE L, ROUGE W, ROUGE SU4 |
| M2 | Precision, Recall, F-measure, Relative utility |

Based on the characterization of text summarization process as shown in Figure 2, classifications of text summarization approaches as shown in Figure 3, databases as discussed in Section 3 and performance evaluation matrices as shown in Figure 4, we bring a consolidated summary of previous studies in text summarization and is shown in Table 4. It discusses the approaches that are used for text summarization; experiment performed using single or multiple documents, types of summary usage, characteristics of the summary and metrics used. The details of the parameters are given in Table 5.

## 7.  ISSUES AND CHALLENGES OCCURS IN TEXT SUMMARIZATION

In the area of text summarization, there are following research issues and challenges occurs during implementation.

### 7.1  Research Issues

- In the case of multi-document text summarization, several issues occurs frequently while evaluation of summary such as redundancy, temporal dimension, co-reference or sentence ordering, etc. which makes very difficult to achieve quality summary. Some other issues occurs such as grammaticality, cohesion, coherence which is harmful for summary.

- The quality of summaries are varying from system to system or person to person. Some person feels some

set of sentences are important for summary, at the same time other person feel the other set of sentences are important for required summary.

### 7.2  Implementation Challenges

- To get the quality summary, quality keywords are required for text summarization.

- There is no standard to identify quality keywords within or multiple documents. The extracted keywords are varying for applying different approaches of keyword extraction.

- Multi-lingual text summarisation is another challenging task.

## 8.  CONCLUSION AND FUTURE DIRECTION

Text summarization is very helpful for users to extract only needed information in stipulated time. In this area, considerable amount of work has been done in the recent past. Due to lack of information and standardization lot of research overlap is a common phenomenon. Since 2012, exhaustive review paper is not published on automatic keyword extraction and text summarization especially in Indian context. Therefore, we thought that, the survey paper covering recent work in keyword extraction and text summarization may ignite the research community for filling some important research gaps. This paper contains the literature review of recent work in text summarization from the point of views of automatic keyword extraction, text databases, summarization process, summarization methodologies and evaluation matrices. Some important research issues in the area of text summarization are also highlighted in the paper.

In future, one can target following direction in the field of summarization:

- Text summarization in low resourced languages especially in Indian language context such as Telugu, Hindi, Tamil, Bengali, etc.
- This work can also be extended to multi-lingual text summarization.
- Multimedia summarization.
- Multi-lingual multimedia summarization.

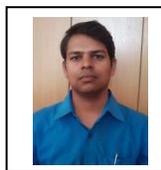

**Santosh Kumar Bharti** is currently pursuing his Ph.D. in CSE from National Institute of Technology Rourkela, India. His research interest includes opinion mining and sarcasm sentiment detection resume. Email-id: sbharti1984@gmail.com

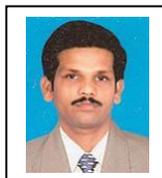

**Korra Sathya Babu** is working as an Assistant Professor in the Dept. of CSE, National Institute of Technology Rourkela India. His research interest includes Data engineering, Data privacy, Opinion mining and Sarcasm sentiment detection. Email-id: prof.ksb@gmail.com

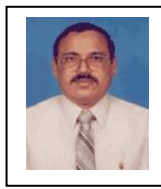

**Sanjay Kumar Jena** is working as a Professor in the Dept. of CSE, National Institute of Technology Rourkela India. His research interest includes Data engineering, Network Security, Opinion mining and Sarcasm sentiment detection. Email-id: skjena@nitrkl.ac.in